\RequirePackage{iftex}
\RequireXeTeX
\documentclass[conference,10pt,letterpaper]{IEEEtran}

\usepackage{cite}
\usepackage{graphicx}
\usepackage{flafter}
\usepackage{amsmath,amssymb}
\usepackage{fontspec}
\usepackage{booktabs}
\usepackage{pgfplots}
\pgfplotsset{compat=1.18}

\definecolor{cdfmblue}{HTML}{496A96}
\definecolor{kappateal}{HTML}{4F918B}
\definecolor{jevorange}{HTML}{D66A23}

\usepackage[hidelinks]{hyperref}
\begin{document}

\title{CausalGrade: Evidence-Guided Causal Graph Scoring with Jev}

% Individual author blocks follow the supplied IEEE conference template.
\author{
\IEEEauthorblockN{Jiangming Zhong}
\IEEEauthorblockA{\textit{Guangdong University of Technology}\\
Guangzhou, China\\
zhongjiangming@mails.gdut.edu.cn}
\and
\IEEEauthorblockN{Jiawei Zhu}
\IEEEauthorblockA{\textit{Guangdong University of Technology}\\
Guangzhou, China\\
heyflyingpig@gmail.com}
\and
\IEEEauthorblockN{Guanting Wu}
\IEEEauthorblockA{\textit{Guangdong University of Technology}\\
Guangzhou, China\\
wuguanting@mails.gdut.edu.cn}
\and
\IEEEauthorblockN{Qiaoxi Cai}
\IEEEauthorblockA{\textit{Guangdong University of Technology}\\
Guangzhou, China\\
caiqiaoxi@mails.gdut.edu.cn}
\and
\IEEEauthorblockN{Wanyi Chen}
\IEEEauthorblockA{\textit{Guangdong University of Technology}\\
Guangzhou, China\\
chenwanyi@mails.gdut.edu.cn}
\and
\IEEEauthorblockN{Jiale Li}
\IEEEauthorblockA{\textit{Guangdong University of Technology}\\
Guangzhou, China\\
lijiale2664@mails.gdut.edu.cn}
}

\maketitle

% Queue the two-column figure on page 1 so it appears at the top of page 2.
\begin{figure*}[!t]
\centering
\includegraphics[width=\textwidth]{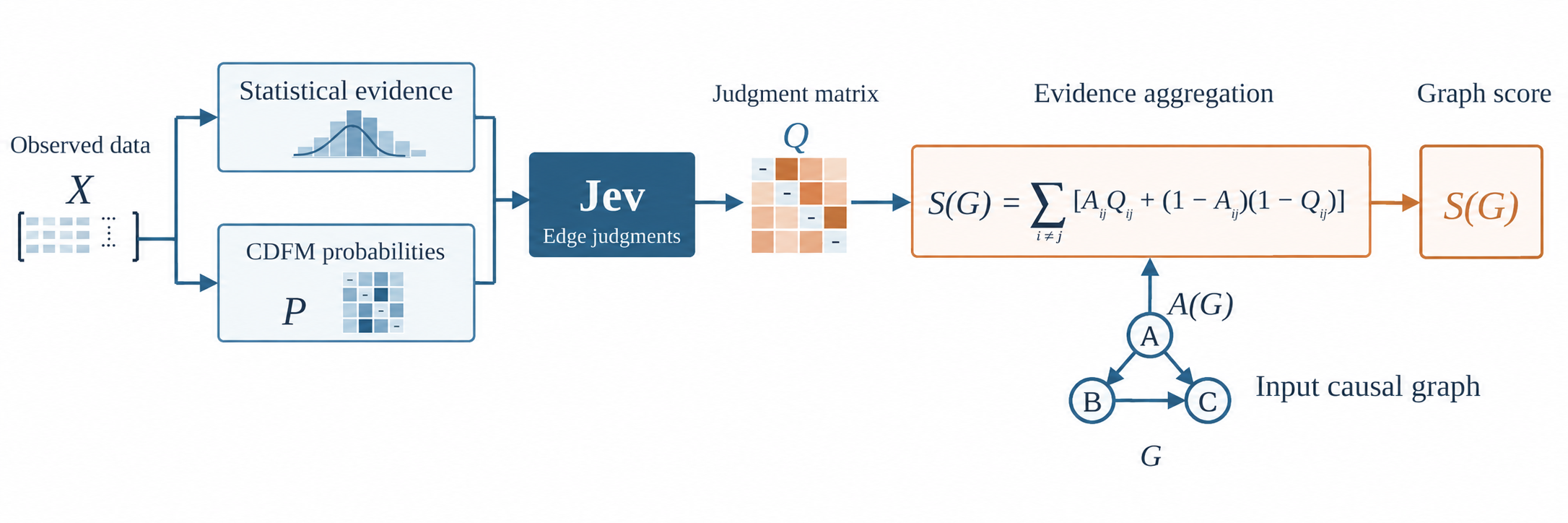}
\caption{Workflow for causal graph evaluation.}
\label{fig:evidence_pipeline}
\end{figure*}

\begin{abstract}

Evaluating the structural quality of candidate causal graphs when the true causal structure is unknown is an important problem in applying causal discovery results.
This paper proposes CausalGrade, a causal graph scoring method based on the probabilistic judgment model Jev.
The method jointly provides Jev with evidence of association, conditional dependence, and directionality from observational data and edge probabilities produced by the causal discovery foundation model CDFM.
Jev produces probabilistic judgments about direct causal edges, and a graph-level score is obtained by measuring the agreement between the adjacency structure of the graph under evaluation and these edge-level judgments.
Across 300 synthetic datasets, the score exhibits a relatively strong negative correlation with structural error in all 20 combinations of four causal discovery algorithms and five variable counts, indicating that higher scores generally correspond to smaller structural errors.
In 500 synthetic graph selection tasks, CausalGrade achieves a best-candidate hit rate of 54.4\%, exceeding the 48.4\% achieved by the Self-Compatibility metric \(\kappa_G\), further supporting its ability to discriminate the structural quality of graphs under evaluation.

\end{abstract}


\section{Introduction}
\label{sec:introduction}

Causal graphs represent direct causal relationships between variables through directed edges and provide an important basis for analyzing intervention effects and explaining system mechanisms.
In practical applications, however, the true causal structure is usually unavailable, and researchers often have to construct causal graphs using observational data or domain knowledge.
Structural errors may affect subsequent causal inference, making it necessary to evaluate a graph's structural quality after it has been obtained.
However, metrics such as structural Hamming distance (SHD) require the ground-truth graph and cannot be directly applied when the true structure is unknown~\cite{eulig2025falsifying,faller2024selfcompatibility}.
This paper addresses the following question: given observational data \(D\) and a graph \(G\) under evaluation, how can we construct a score that indicates structural quality and supports candidate comparison?

Existing studies provide different bases for evaluation without ground truth by examining graph--data compatibility and consistency among algorithm outputs.
DAGitty~\cite{textor2016dagitty} tests the conditional independence relations implied by a graph to identify structural constraints that are inconsistent with observational data.
Eulig et al.~\cite{eulig2025falsifying} further use node permutations to construct a reference, testing whether a given graph is more consistent with the independence relations in the data than randomly permuted graphs.
Self-Compatibility~\cite{faller2024selfcompatibility} examines whether the causal structures or intervention effects obtained by the same algorithm on different variable subsets are consistent, and demonstrates the association between this consistency and structural error, as well as its value for model selection.

These studies show that graphs can be evaluated using references other than the true structure.
We further consider organizing multiple types of evidence from the current data as a basis for judgments about direct causal edges.
A challenge in this design is that different statistical evidence may provide conflicting support for direct edges and their directions.
It is therefore necessary to distinguish the meanings of different statistics, organize heterogeneous evidence into probabilistic judgments about direct causal edges, and aggregate these edge-level judgments for graph-level evaluation.

To this end, we propose CausalGrade, whose overall workflow is shown in Fig.~\ref{fig:evidence_pipeline}.
To convert shared evidence into edge-level probabilistic judgments, we use the Jev model provided by TypeSafe~\cite{typesafejev}, through its interface that returns probabilities for yes/no questions given a context.
We construct a statistical report \(T(D)\) from \(D\), summarizing cues about association, conditional dependence, and directionality, and incorporate the edge probability matrix \(P\) produced by CDFM~\cite{qiao2026cdfm}.
Jev then produces a judgment matrix \(Q=(Q_{ij})\) for direct-edge propositions concerning each ordered variable pair.
We subsequently aggregate \(Q\) with the adjacency matrix \(A(G)\) of the graph under evaluation to obtain the score \(S(G)\).
The score measures the agreement between a candidate structure and the edge-level judgments; its effectiveness as a proxy for structural quality is examined through its association with structural error and its performance in candidate selection.
Graphs evaluated on the same data share one edge-level judgment matrix \(Q\), allowing them to be evaluated on a common evidential basis.

The main contributions of this paper are as follows:
\begin{enumerate}
\item We propose CausalGrade, which organizes a statistical report and CDFM edge probabilities as evidence for direct-edge judgments and uses Jev to generate an edge-level probabilistic judgment matrix shared by different graphs evaluated on the same data.

\item We construct a graph score that accounts for both edge presence and edge absence, and establish its algebraic correspondence with SHD2 estimated from the edge-level judgments, where SHD2 denotes structural Hamming distance with edge reversals counted twice.

\item We examine the effectiveness of the score through evaluation of fixed algorithm-generated graphs, graph selection on synthetic data and real-world systems, and input ablation.
In 500 synthetic graph selection tasks, CausalGrade achieves a best-candidate hit rate of 54.4\%, exceeding the 19.6\% achieved by directly aggregating CDFM edge probabilities and the 48.4\% achieved by \(\kappa_G\).
Compared with \(\kappa_G\), it reduces the mean SHD2 gap between the selected graph and the best candidate within each task by 28.8\%.
\end{enumerate}

\section{Related Work}
\label{sec:related_work}

\subsection{Structure Evaluation without Ground Truth}
The \(\kappa_G\) metric in Self-Compatibility~\cite{faller2024selfcompatibility} takes the latent projection of the graph on the full variable set onto each variable subset, computes its SHD from the graph relearned by the same algorithm on the corresponding subset data, and averages these distances.
Latent projection accounts for indirect paths and latent confounding induced by omitted variables.
The reference for this evaluation comes from structural outputs across variable sets.
In our method, direct-edge judgments \(Q\) are constructed on the full variable set, and their agreement with the graph under evaluation provides the basis for evaluation.

\subsection{Pretrained Causal Discovery}
CDFM~\cite{qiao2026cdfm} is pretrained on data generated by diverse structural causal models and their known graphs to learn a mapping from observational data to causal structure.
Its training distribution covers different graph structures, functional mechanisms, and noise types.
The model extracts representations through column embeddings and alternating attention along the sample and variable dimensions, then constructs separate cause and effect representations.
Asymmetric bilinear decoding produces directed-edge scores, which are transformed by a sigmoid function into an edge probability matrix \(P\).
These probabilities reflect predictions on the current data based on structural information learned during pretraining, providing a model-based reference distinct from statistical tests performed on the data.

\subsection{Statistical Evidence and Probabilistic Judgment}
KCI~\cite{zhang2011kci} uses kernel methods to test independence given conditioning variables, whereas research on additive noise models~\cite{hoyer2008anm} exploits independence between cause and noise to identify direction under the corresponding model assumptions.
These two lines of work support our construction of conditional dependence diagnostics and directional evidence, respectively.
Test \(p\)-values describe statistical significance and play a different role from probabilistic judgments about edge presence.
Regarding probabilistic judgment, Deu\ss er et al.~\cite{deusser2026jev} evaluate Jev on tasks including classification, reasoning, and scoring.
We apply this judgment interface to direct-edge propositions supported by statistical evidence and examine its effectiveness for causal graph evaluation in our experiments.

% =========================================================
% Method
% =========================================================

\section{Method}
\label{sec:method}

\subsection{Problem Formulation}
\label{subsec:problem_formulation}

Given observational data \(D \in \mathbb{R}^{n \times d}\) containing \(n\) samples and \(d\) variables, together with a directed acyclic graph \(G\) on the same variable set, we use a statistical report \(T(D)\) and the CDFM~\cite{qiao2026cdfm} edge probability matrix \(P\) to provide a basis for scoring the structural quality of \(G\) when the true structure is unknown.
In the adjacency matrix \(A(G) \in \{0,1\}^{d \times d}\), \(A_{ij}=1\) indicates \(X_i \to X_j\), and the entry is zero otherwise.

CausalGrade first uses the statistical report \(T(D)\) and the CDFM edge probability matrix \(P\) to generate an edge-level probabilistic judgment matrix \(Q\), then aggregates \(Q\) with \(A(G)\) to obtain a graph score \(S(G)\).
A higher score indicates greater agreement between the adjacency structure of the graph under evaluation and the edge-level judgments.

\subsection{Evidence Construction and Jev Judgment}
\label{subsec:evidence_judgment}

The statistical report \(T(D)\) first summarizes sample size, missing-value rate, mean, standard deviation, quantiles, skewness, and kurtosis to describe data quality and distributional characteristics.
It then organizes evidence at three levels: association, conditional dependence, and directional asymmetry.

\paragraph{Association Evidence}
The report includes Pearson correlation, Spearman rank correlation, and normalized RBF kernel alignment, which characterize linear correlation, monotonic correlation, and statistical dependence that may include nonlinear relationships, respectively.

\paragraph{Conditional Dependence Evidence}
For linear conditional dependence, the report summarizes the minimum absolute partial correlation with up to two conditioning variables, the corresponding conditioning set, the Fisher--Z test, the number of conditioning sets for which independence is not rejected, and the partial correlation after controlling for all remaining variables.

Nonlinear conditional dependence is assessed using the kernel-based conditional independence test (KCI~\cite{zhang2011kci}), which tests \(X_i \perp X_j \mid Z\).
For each candidate conditioning variable, we compute its kernel association strength with \(X_i\) and with \(X_j\), and use the smaller of the two values as its ranking score.
Candidate variables are ranked in descending order of this score, and conditioning sets \(Z\) are formed using the top-ranked variable, the two top-ranked variables, and all remaining variables, respectively.
KCI uses Gaussian kernels to account for the conditioning variables.
The report records the conditioning sets, \(p\)-values, normalized residual dependence strength, and significance categories.
Together, the test results across conditioning sets describe changes in dependence between variables and provide conditional dependence evidence for direct-edge judgments.

\paragraph{Directional Evidence}
We perform prediction in both directions using nonlinear kernel regression with five-fold cross-fitting, recording out-of-fold \(R^2\), Spearman correlation and normalized HSIC between inputs and residuals, and comparing residual dependence in the reverse and forward directions.
Under an applicable additive noise model~\cite{hoyer2008anm}, the cause is independent of the noise; weaker dependence between inputs and residuals therefore supports the corresponding direction.
We further let \(Z\) contain all variables other than \(X_i\) and \(X_j\), perform five-fold cross-fitting to predict \(X_j\) using either \(Z\) alone or both \(Z,X_i\), and compare their out-of-fold mean squared errors:
\begin{equation}
\label{eq:incremental_gain}
\Delta_{i \to j}
=
\frac{
\operatorname{MSE}(X_j \mid Z)
-
\operatorname{MSE}(X_j \mid Z,X_i)
}{
\operatorname{MSE}(X_j \mid Z)
}.
\end{equation}
A positive gain indicates that adding the source variable provides additional predictive information.
With out-of-fold predictions held fixed, we resample the sample-level errors from the two prediction settings in pairs to estimate the variability of the gain.
The report also includes the proportion of cross-validation folds with positive gain and the variation in association and directional measures across sample subsets to describe the stability of the evidence.

\paragraph{Model Evidence and Jev Judgment}
CDFM~\cite{qiao2026cdfm} generates \(P=(p_{ij})\) from the same data, where \(p_{ij}\) denotes the model's probability output for the directed edge \(X_i\to X_j\).
The statistical report provides dependence diagnostics for the current data, whereas \(P\) provides structural predictions from the pretrained model for these data.

We denote variables by \(X_1,\ldots,X_d\) in the column order of the observational data, establishing a one-to-one correspondence between each variable pair \((X_i,X_j)\) in the statistical report and the edge probability \(p_{ij}\) in the CDFM matrix.
We then jointly provide Jev with the statistical report and the full probability matrix and ask, for each ordered variable pair, ``Is \(X_i\) a direct cause of \(X_j\)?''
Let \(\mathcal{J}_{ij}\) denote Jev's probabilistic judgment process for the presence of the directed edge \(X_i\to X_j\), yielding
\begin{equation}
\label{eq:jev_judgment}
Q_{ij}
=
\mathcal{J}_{ij}\!\left(T(D),P\right)
\in [0,1],
\qquad i \ne j.
\end{equation}

The graph \(G\) under evaluation does not enter this judgment process.
The direct-edge question is posed through Jev's Noul probability interface~\cite{typesafenoul}, which returns a number in \([0,1]\) representing the probability of a ``yes'' answer to the direct-edge proposition.
We denote this edge-level judgment by \(Q_{ij}\).

\subsection{Graph Scoring and Evidence Aggregation}
\label{subsec:graph_scoring}

We convert the graph \(G\) under evaluation into its adjacency matrix \(A(G)\) and align it with the judgment matrix \(Q\) according to the data column order.
Let \(A_{ij}\) denote the corresponding entry of \(A(G)\). We define
\begin{equation}
\label{eq:graph_score}
S(G)
=
\sum_{i \ne j}
\bigl[
A_{ij}Q_{ij}
+
(1-A_{ij})(1-Q_{ij})
\bigr].
\end{equation}
When the graph contains an edge, i.e., \(A_{ij}=1\), the contribution of that entry is \(Q_{ij}\); otherwise, it is \(1-Q_{ij}\).
Taking \(Q_{ij}\) as an estimate of the probability that the true edge is present and summing over ordered variable pairs gives the corresponding estimate of structural error:
\begin{equation}
\label{eq:estimated_structural_error}
\begin{aligned}
\widehat{L}_Q(G)
&=
\sum_{i \ne j}
\bigl[
A_{ij}(1-Q_{ij})
+
(1-A_{ij})Q_{ij}
\bigr] \\
&= d(d-1)-S(G).
\end{aligned}
\end{equation}

By linearity of expectation, the score represents the expected number of correct adjacency entries computed from the edge-level probabilistic judgments, accounting for both edge presence and edge absence.
We use SHD2 to denote structural Hamming distance with edge additions and deletions each counted once and edge reversals counted twice, measuring the structural difference between the graph under evaluation and the reference graph.
The quantity \(\widehat{L}_Q(G)\) corresponds to SHD2 estimated from \(Q\).
For a fixed variable set, maximizing \(S(G)\) is equivalent to minimizing \(\widehat{L}_Q(G)\).

% =========================================================
% Experiments and Results
% =========================================================

\section{Experiments and Results}
\label{sec:experiments_results}

\subsection{Experimental Setup}
\label{subsec:setup}

We evaluate the graph scoring method in four settings: evaluation of fixed algorithm-generated graphs, graph selection on synthetic data, graph selection on real-world data, and input ablation.
Synthetic data are constructed following the data-generating mechanisms described in the CDFM paper~\cite{qiao2026cdfm}.
We first construct a directed acyclic graph (DAG) for the specified variable count, then generate observational data under the specified mechanism and noise settings, retaining the constructed DAG as the ground truth for evaluation.
We use 15 mechanisms to generate data from graphs: CAM, CPT, discrete ANM, and PNL; standard, heteroscedastic, and ordered variants of both Linear and RFF; Physical and its heteroscedastic variant; measurement error, rounding, and time lag.
Jev version 1.13.0 is fixed across all experiments.
The CDFM score is obtained from the CDFM edge probability matrix \(P\):
\begin{equation}
\label{eq:cdfm_score}
S_{\mathrm{CDFM}}(A)
=
\sum_{i\ne j}
\left[
A_{ij}P_{ij}+(1-A_{ij})(1-P_{ij})
\right].
\end{equation}

The score correlation experiment uses 300 datasets, covering all combinations of five variable counts (8, 10, 15, 20, and 30), four sample sizes (500, 1000, 2000, and 3000), and 15 data-generating mechanisms.
For each dataset, PC, GES, DirectLiNGAM, and NOTEARS are run separately to generate causal graphs.
We compute the CDFM score, the CausalGrade score, and SHD2 relative to the true structure for each graph.
Within each algorithm, we compute the Spearman rank correlation between each score and SHD2.
A correlation closer to \(-1\) indicates a stronger inverse rank association between the score and structural error: graphs with higher scores generally have lower SHD2, indicating that the score is more informative about graph quality.

Synthetic graph selection uses 500 datasets, each with 10 variables and 1000 samples, covering the 15 mechanisms above and six graph families: random graphs (ER), scale-free graphs (SF), scale-free graphs with triadic closure (SFT), small-world graphs (WS), stochastic block model graphs (SBM), and random geometric graphs (GRG).
We compare CDFM, the \(\kappa_G\) metric from Self-Compatibility~\cite{faller2024selfcompatibility}, and CausalGrade.
Graph selection on real-world data uses Light from a physical light-tunnel system~\cite{gamella2025causalchambers} and the Sachs cell-signaling dataset~\cite{sachs2005}.
Light is obtained from the CDFM repository and contains 10000 rows and 20 variables.
Sachs is obtained from the cd3cd28 condition in SEMgraph and contains single-cell flow cytometry measurements with 853 rows and 11 variables.
We draw 400 and 200 samples from the two datasets, respectively, each containing 800 rows, and use each dataset's reference structure as ground truth in the subsequent evaluation.

In both graph selection experiments, for each dataset, we apply thresholds of 0.1, 0.3, 0.5, 0.7, and 0.9, retaining only directed edges whose CDFM-predicted probabilities exceed the threshold.
After removing cycles according to a fixed rule, this produces 5 graphs per dataset for evaluation.
All methods share these graphs and the same tie-breaking order.
This design compares scoring methods while fixing the source of the candidates, but the thresholds change both edge counts and connectivity.
The selection results therefore do not yet separate the contribution of sparsity selection from that of discriminating specific connections and directions.
CDFM and CausalGrade select the graph with the highest score, whereas \(\kappa_G\) selects the graph with the smallest structural discrepancy across subsets.
For each instance, \(\kappa_G\) uses 40 variable subsets and compares structures inferred on the subsets with the corresponding marginalized structures of the full candidate graph.
One edge-level judgment matrix \(Q\) is generated per dataset and held fixed for scoring all graphs evaluated on that dataset.
The hit rate is the proportion of tasks in which a method selects a graph with the minimum SHD2 among the 5 graphs in the same candidate set; any tied best candidate counts as a hit.
Selection loss is the SHD2 of the selected graph minus the minimum SHD2 in that set, and is zero for a hit.
These metrics respectively reflect how often the best candidate within a task is selected and the additional structural error relative to the minimum SHD2.
The hit rate for each dataset is computed using its reference structure and candidate set, giving selection success rates a consistent meaning across variable counts.
The reference structure for the current evaluation task is not used in evidence construction, Jev judgment, or candidate graph selection.
Only after all predictions are frozen does an independent evaluation procedure access the reference structure and compute SHD2, hit rate, and selection loss.

The ablation reuses the 500 synthetic datasets and their graphs under evaluation, changing only the inputs to Jev.

\subsection{Association with Graph Quality}
\label{subsec:association}

This experiment examines whether the score can indicate the structural quality of a fixed graph.
The four algorithms represent approaches to graph construction based on conditional independence testing, score-based search, linear non-Gaussian modeling, and continuous optimization, respectively.
We hold each algorithm's outputs and their corresponding SHD2 values fixed and compare the scores obtained from CDFM probabilities and Jev judgments.
When pooling across variable counts, both scores and SHD2 are divided by \(d(d-1)\), where \(d\) is the number of variables, before computing correlations within each algorithm to ensure a common comparison scale.
The results are reported in Table~\ref{tab:quality_association}.

\begin{table}[htbp]
\caption{Spearman Rank Correlations between Scores and SHD2 in the Score Correlation Experiment}
\label{tab:quality_association}
\centering
\footnotesize
\begin{tabular}{lrr}
\toprule
Algorithm & CDFM:$\rho$ & CausalGrade:$\rho$ \\
\midrule
PC           & $-0.534$ & $-0.763$ \\
GES          & $-0.558$ & $-0.704$ \\
DirectLiNGAM & $-0.512$ & $-0.523$ \\
NOTEARS      & $-0.490$ & $-0.762$ \\
\bottomrule
\end{tabular}
\end{table}

Table~\ref{tab:quality_association} shows that CausalGrade exhibits a relatively strong negative correlation with SHD2 for all four algorithms, indicating that higher scores generally correspond to graphs closer to the reference structure.
The correlations for PC, GES, DirectLiNGAM, and NOTEARS are \(-0.763\), \(-0.704\), \(-0.523\), and \(-0.762\), respectively, all stronger than those of CDFM.
Further stratification by algorithm and variable count shows relatively strong negative correlations in all 20 combinations of 4 algorithms and 5 variable counts, with stronger correlations than CDFM in 14 combinations.
Together, the pooled and stratified results support the score's role as a proxy for structural quality for the algorithm outputs examined.

\subsection{Candidate Graph Selection}
\label{subsec:selection}

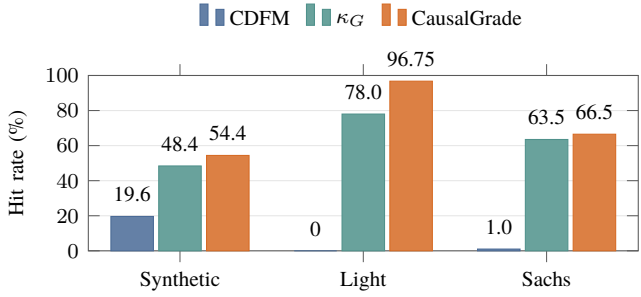
\begin{figure}[htbp]
\centering
\begin{tikzpicture}
\begin{axis}[
width=\columnwidth,
height=3.9cm,
ybar,
bar width=16pt,
ymin=0,
ymax=100,
xmin=0.5,
xmax=3.5,
xtick={1,2,3},
xticklabels={Synthetic,Light,Sachs},
ytick={0,20,40,60,80,100},
ylabel={Hit rate (\%)},
tick label style={font=\footnotesize},
label style={font=\footnotesize},
ymajorgrids=true,
grid style={gray!20},
axis line style={black!65},
tick style={black!65},
legend style={
at={(0.5,1.20)},
anchor=south,
legend columns=3,
draw=none,
font=\footnotesize,
/tikz/every even column/.append style={column sep=4pt}
},
point meta=explicit symbolic,
nodes near coords,
every node near coord/.append style={
font=\footnotesize,
text=black,
yshift=2pt
},
clip=false
]
\addplot+[
draw=cdfmblue,
fill=cdfmblue!85
] coordinates {
(1,19.6) [19.6]
(2,0)    [0]
(3,1.0)  [1.0]
};

\addplot+[
draw=kappateal,
fill=kappateal!85
] coordinates {
(1,48.4) [48.4]
(2,78.0) [78.0]
(3,63.5) [63.5]
};

\addplot+[
draw=jevorange,
fill=jevorange!85
] coordinates {
(1,54.4)  [54.4]
(2,96.75) [96.75]
(3,66.5)  [66.5]
};

\legend{CDFM,$\kappa_G$,CausalGrade}
\end{axis}
\end{tikzpicture}
\caption{Best-candidate hit rates across three types of tasks.}
\label{fig:selection_hit}
\end{figure}

Having established the association with graph quality, we further examine how the score helps with graph selection in practice.
Fig.~\ref{fig:selection_hit} compares the best-candidate hit rates of the three methods within the same candidate sets.

Across the 500 synthetic tasks, CausalGrade records 272 hits, achieving a hit rate of 54.4\%.
This exceeds the 19.6\% achieved by CDFM and the 48.4\% achieved by \(\kappa_G\) by 34.8 and 6.0 percentage points, respectively.
Mean selection loss decreases from 4.532 for CDFM and 2.504 for \(\kappa_G\) to 1.784, corresponding to reductions of 60.6\% and 28.8\%, respectively.
In instance-by-instance comparisons, CausalGrade selects graphs closer to the true structure than CDFM in 340 tasks and outperforms \(\kappa_G\) in 155 tasks.

These improvements in hit rate and selection loss show that, within the candidate sets examined, CausalGrade both selects the best graph more often and reduces the mean additional structural error relative to the best candidate.

For the real-world systems, CausalGrade records 387 hits in the 400 Light tasks, achieving a hit rate of 96.75\%, compared with 0\% for CDFM and 78.0\% for \(\kappa_G\).
This represents an improvement of 18.75 percentage points over \(\kappa_G\).
In the 200 Sachs tasks, CausalGrade records 133 hits, achieving 66.5\%, compared with 1.0\% for CDFM and 63.5\% for \(\kappa_G\).

Across all three types of graph selection tasks---synthetic data, Light, and Sachs---CausalGrade achieves the highest hit rate among the methods compared, supporting its use for graph selection within the candidate sets examined.

\subsection{Input Ablation}
\label{subsec:ablation}

To distinguish the contributions of the two types of input, we hold the data, candidates, Jev model, direct-edge question, scoring rule, and tie-breaking rule fixed, removing only the statistical report \(T(D)\) or the CDFM edge probability matrix \(P\) in each ablation.
With the full input, mean selection loss is 1.784 and the hit rate is 54.4\%.
Removing the statistical report gives 6.140 and 18.8\%, respectively; removing the CDFM probabilities gives 2.102 and 53.8\%.
Compared with the two ablations, the full method reduces mean loss by 4.356 and 0.318, respectively.
Using 10000 bootstrap resamples stratified by mechanism and paired by dataset instance, the 95\% intervals for these loss reductions are [3.778,4.936] and [0.108,0.538], respectively.
The results indicate that the statistical report contributes the main gains in selection performance, while CDFM probabilities provide additional evidence for judgment.
Table~\ref{tab:input_ablation} summarizes the results for the three input settings.

\begin{table}[htbp]
\caption{Input Ablation Results (500 Synthetic Datasets)}
\label{tab:input_ablation}
\centering
\footnotesize
\begin{tabular}{lcc}
\toprule
Input setting & Mean selection loss$\downarrow$ & Hit rate (\%)$\uparrow$ \\
\midrule
Full input     & \textbf{1.784} & \textbf{54.4} \\
Without statistical report & 6.140          & 18.8          \\
Without CDFM probabilities & 2.102          & 53.8          \\
\bottomrule
\end{tabular}
\end{table}

\section{Discussion}
\label{sec:discussion}

The experiments show that combining statistical evidence from the current data with model predictions provides an effective basis for evaluating the quality of the causal graphs examined.
The ablations nevertheless reveal an asymmetry in the contributions of the two types of evidence: adding the statistical report while retaining CDFM probabilities reduces mean selection loss by 4.356, whereas adding CDFM probabilities while retaining the statistical report reduces it by 0.318.
Understanding this scoring performance requires a closer examination of the information that the two types of evidence provide to Jev.
CDFM probabilities \(P\) summarize the model's directed-edge predictions for the current data, whereas the statistical report retains diagnostic details, including changes in conditional dependence across conditioning sets, differences between fits in the two directions, and incremental prediction, providing an explicit basis for direct-edge judgments.
One possible explanation is that these explicit diagnostics give Jev an additional basis for assessing model predictions and forming edge-level judgments in light of the current data.
However, the current ablations cannot determine which diagnostics play the main role, nor can they distinguish whether the gains arise from the additional diagnostic information or from how Jev uses the organization of its inputs.

Beyond the effects of the two types of evidence on edge-level judgments, the aggregation used in graph-level scoring also affects candidate graph selection, with a preference for sparse structures as one direct consequence.
According to~\eqref{eq:graph_score}, removing an edge increases the score when \(Q_{ij}<0.5\).
In our experiments, the graphs under evaluation are constructed from the same CDFM output using different thresholds, so edge counts and connectivity vary together with the threshold.
The current experiments have not separated the contribution of sparsity selection from that of discriminating specific connections and directions.
Insufficient evidence in finite samples may therefore lead to the removal of true edges, with part of the improvement in structural error arising from sparsification.
Future work could compare candidates with equal edge counts but different connections or directions across graph densities and sample sizes to further clarify the score's ability to discriminate structure and retain weak causal edges.

\section{Conclusion}
\label{sec:conclusion}

This paper proposes CausalGrade, which uses Jev to convert statistical reports and CDFM edge probabilities into probabilistic judgments about direct edges, then constructs a graph-level score from their agreement with the adjacency structure of the graph under evaluation, accounting for both edge presence and edge absence.
Under the settings examined, the score indicates structural quality and supports candidate comparison.
By separating graph quality evaluation from the specific graph construction process, CausalGrade provides a reusable scoring method for causal graph evaluation when the true structure is unknown.

\end{document}